\documentclass[11pt,letterpaper]{article}
\usepackage{emnlp2016}
\usepackage{times}
\usepackage{latexsym}
\usepackage{hyperref}
\usepackage{url}
\usepackage{amsfonts}
\usepackage{graphicx}
\usepackage{amsmath}
\usepackage{booktabs}
\usepackage{multirow}
\usepackage{bbm}

\DeclareMathOperator*{\argmax}{argmax}

 \emnlpfinalcopy

\title{Sequence-Level Knowledge Distillation}
\author{Yoon Kim \\ {\tt yoonkim@seas.harvard.edu} 
\And 
Alexander M. Rush \\ {\tt srush@seas.harvard.edu} 
\AND 
\textnormal{School of Engineering and Applied Sciences} \\
\textnormal{Harvard University} \\
Cambridge, MA, USA  \\
}
\date{}

\begin{document}

\maketitle

\newcommand{\fix}{\marginpar{FIX}}
\newcommand{\new}{\marginpar{NEW}}
\newcommand{\xvec}{\mathbf{x}}
\newcommand{\yvec}{\mathbf{y}}
\newcommand{\cvec}{\mathbf{c}}
\newcommand{\zvec}{\mathbf{z}}
\newcommand{\svec}{\mathbf{s}}
\newcommand{\tvec}{\mathbf{t}}
\newcommand{\mcL}{\mathcal{L}}
\newcommand{\mcT}{\mathcal{T}}
\newcommand{\mcY}{\mathcal{Y}}
\newcommand{\mcV}{\mathcal{V}}
\newcommand{\mcC}{\mathcal{C}}
\newcommand{\mcA}{\mathcal{A}}
\newcommand{\context}{\mathbf{y}_{\mathrm{c}}}
\newcommand{\embcontext}{\mathbf{\tilde{y}}_{\mathrm{c}}}
\newcommand{\inpcontext}{\mathbf{\tilde{x}}}
\newcommand{\start}{\mathbf{\tilde{y}}_{\mathrm{c0}}}
\newcommand{\End}{\mathrm{\texttt{</s>}}}

\newcommand{\Uvec}{\mathbf{U}}
\newcommand{\Evec}{\mathbf{E}}
\newcommand{\Gvec}{\mathbf{G}}
\newcommand{\Fvec}{\mathbf{F}}
\newcommand{\Pvec}{\mathbf{P}}
\newcommand{\pvec}{\mathbf{p}}
\newcommand{\Vvec}{\mathbf{V}}
\newcommand{\Wvec}{\mathbf{W}}
\newcommand{\hvec}{\mathbf{h}}
\newcommand{\wvec}{\mathbf{w}}
\newcommand{\uvec}{\mathbf{u}}
\newcommand{\vvec}{\mathbf{v}}
\newcommand{\bvec}{\mathbf{b}}
\newcommand{\reals}{\mathbb{R}}
\newcommand\given{\,|\,}

\pdfinfo{
/Title (Sequence-Level Knowledge Distillation)
/Author (Yoon Kim, Alexander M. Rush)}
\maketitle

\begin{abstract} 
  Neural machine translation (NMT) offers a novel alternative
  formulation of translation that is potentially simpler than
  statistical approaches. However to reach competitive performance,
  NMT models need to be exceedingly large.
  In this paper we consider
  applying \textit{knowledge distillation} approaches 
  \cite{Bucila2006,Hinton2015} that have proven
  successful for reducing the size of neural models in other domains
  to the problem of NMT. We demonstrate that standard knowledge
  distillation applied to word-level prediction can be effective for
  NMT, and also introduce two novel \textit{sequence-level} versions
  of knowledge distillation that further improve performance, and
  somewhat surprisingly, seem to eliminate the need for beam search (even when
  applied on the original teacher model). Our best
  student model runs $10$ times faster than its state-of-the-art teacher with
  little loss in performance. 
  It is also significantly better than a baseline model trained without
   knowledge distillation: by $4.2/1.7$ BLEU with greedy decoding/beam search. 
   Applying weight pruning on top of knowledge distillation results in a student model
   that has $13 \times$ fewer parameters than the original teacher model,
   with a decrease of $0.4$ BLEU.
\end{abstract}

\section{Introduction}

Neural machine translation (NMT)
\cite{Kalchbrenner2013,Cho2014,Sutskever2014,Bahdanau2015} is a deep learning-based
method for translation that has recently shown promising results as
an alternative to statistical approaches.  NMT systems directly model
the probability of the next word in the target sentence simply by
conditioning a recurrent neural network on the source sentence and
previously generated target words. 

While both simple and surprisingly accurate, NMT systems typically
need to have very high capacity in order to perform well: \newcite{Sutskever2014} 
used a $4$-layer LSTM with $1000$ hidden
units per layer (herein $4\times1000$) and \newcite{Zhou2016} obtained
state-of-the-art results on English $\rightarrow$ French with a
$16$-layer LSTM with $512$ units per layer. The sheer size of the
models requires cutting-edge hardware for training and makes using the
models on standard setups very challenging. 

This issue of excessively large networks has been observed in several
other domains, with much focus on fully-connected and convolutional
networks for multi-class classification. Researchers have particularly
noted that large networks seem to be necessary for training, but learn
redundant representations in the process \cite{Denil2013}. Therefore compressing deep models into
smaller networks has been an active area of research. As deep learning
systems obtain better results on NLP tasks, compression also becomes
an important practical issue with applications such as running
deep learning models for speech and translation locally on
cell phones.

Existing compression methods generally fall into two categories: (1)
\textit{pruning} and (2) \textit{knowledge distillation}.
\textit{Pruning} methods 
\cite{LeCun1990,He2014,Han2016}, zero-out weights or entire neurons
based on an importance criterion: \newcite{LeCun1990} use (a diagonal approximation to) the
Hessian to identify weights whose removal minimally impacts the
objective function, while \newcite{Han2016} remove
weights based on thresholding their absolute values. 
\textit{Knowledge distillation} approaches
\cite{Bucila2006,Ba2014,Hinton2015} learn
a smaller \textit{student} network to mimic the original
\textit{teacher} network by minimizing the loss (typically $L_2$ or
cross-entropy) between the student and teacher output. 

In this work, we investigate knowledge distillation  in the context of
neural machine translation. We note that NMT differs
from previous work which has mainly explored non-recurrent
models in the multi-class prediction setting. For NMT,
while the model is trained on multi-class prediction at the word-level, it is tasked
with predicting complete sequence outputs conditioned on previous
decisions. With this difference in mind, we experiment with standard knowledge
distillation for NMT and also propose two new versions of the approach
that attempt to approximately match the sequence-level (as opposed to word-level)
distribution of the teacher network. This
sequence-level approximation leads to a simple training procedure wherein the student
network is trained on a newly generated dataset that is the result of running beam search 
with the teacher network.

We run experiments to compress a large
state-of-the-art $4\times1000$ LSTM model, and find that with sequence-level knowledge 
distillation we are
able to learn a $2\times500$ LSTM that roughly matches the performance
of the full system. We see similar results
compressing a $2\times 500$ model down to $2\times 100$ on a smaller
data set.  Furthermore, we observe that  our proposed approach has other
benefits, such as not requiring any beam search at test-time. As a
result we are able to perform greedy decoding on the $2\times500$
model $10$ times faster than beam search on the $4\times1000$ model
with comparable performance. Our student models can even be run
efficiently on a standard smartphone.\footnote{https://github.com/harvardnlp/nmt-android}
Finally, we apply weight pruning on top of the student network to obtain a model that has 
$13\times$ fewer parameters than the original teacher model.
 We have released all the code for the models 
described in this paper.\footnote{https://github.com/harvardnlp/seq2seq-attn}

\section{Background}

\subsection{Sequence-to-Sequence with Attention}\label{NMT}
Let $\mathbf{s} = [s_1, \dots, s_I]$ and $\mathbf{t} = [t_1, \dots, t_J]$ be 
 (random variable sequences representing) the source/target sentence, 
with $I$ and $J$ respectively being the source/target lengths. Machine translation
involves finding the most probable target sentence given the source: 
\begin{equation*}
\argmax_{\tvec \in \mcT} p(\tvec \given \svec)
\end{equation*}
where $\mcT$ is the set of all possible sequences.
NMT models parameterize $p(\tvec\given\svec)$ 
with an \textit{encoder} neural network which reads the source sentence and a \textit{decoder}
neural network which produces a distribution over the target sentence (one word at a time) given the source. 
We employ the attentional architecture from 
\newcite{Luong2015}, which achieved state-of-the-art results on English $\rightarrow$
German translation.\footnote{Specifically, we use the \textit{global-general} attention model
with the \textit{input-feeding} approach. We refer the reader to the original paper for further details.}

\begin{figure*}[htp]\label{fig1}
\centering
\includegraphics[width=16.5cm]{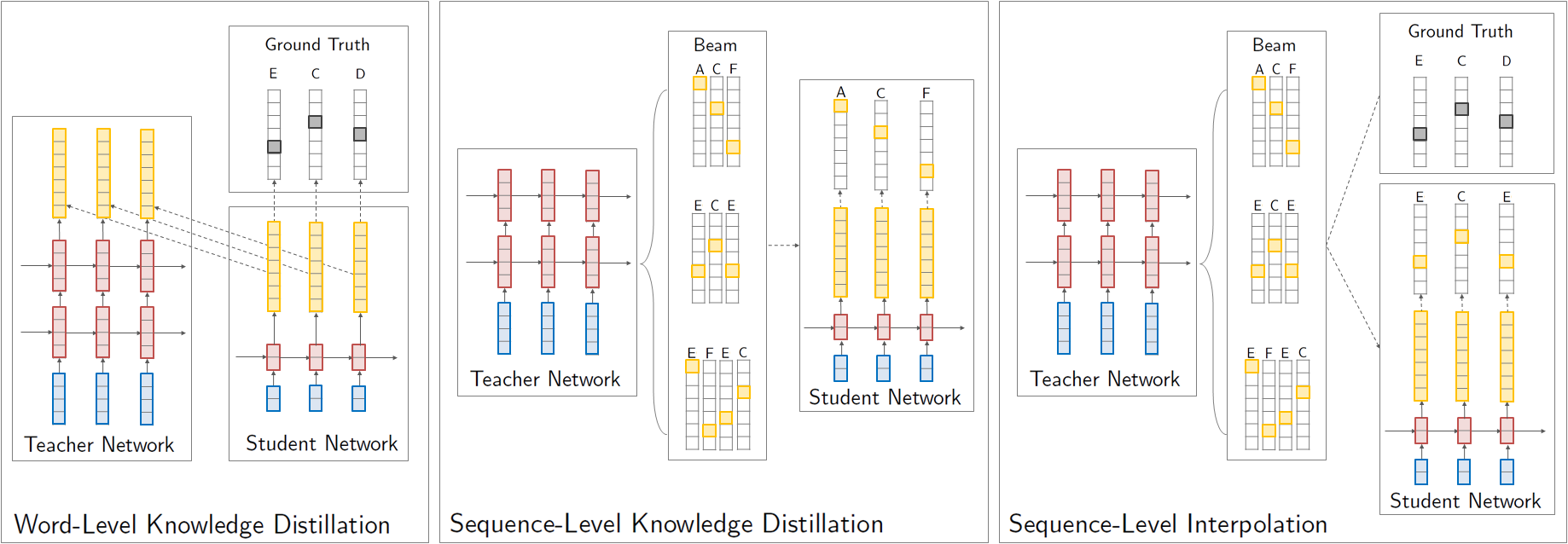}
\caption{Overview of the different knowledge distillation approaches.
In word-level knowledge distillation (left) cross-entropy is minimized between the
  student/teacher distributions (yellow) for each word in the actual target sequence (\textsf{ECD}), as well as
  between the student distribution and the degenerate data distribution, which has all of its
  probabilitiy mass on one word (black). In sequence-level knowledge distillation (center)
  the student network is trained on the output from beam search of the teacher network that
  had the highest score (\textsf{ACF}). In sequence-level interpolation (right) the student is trained on the output
 from beam search of the teacher network that had the highest $sim$ with the target sequence (\textsf{ECE}).}
\end{figure*}
\subsection{Knowledge Distillation}\label{KD}

Knowledge distillation describes a class of methods for training a
smaller \textit{student} network to perform better by learning from a
larger \textit{teacher} network (in addition to learning from the
training data set). We generally assume that the teacher has previously
been trained, and that we are estimating parameters for the student.
Knowledge distillation suggests training by matching the student's
predictions to the teacher's predictions. For classification this
usually means matching the probabilities either via $L_2$ on the
$\log$ scale \cite{Ba2014} or by cross-entropy
\cite{Li2014,Hinton2015}.

Concretely, assume we are learning a multi-class classifier over 
 a data set of examples of the form $(x, y)$ 
with possible classes $\mcV$. The usual training criteria is to minimize NLL for each example 
from the 
training data,  
\begin{equation*}
\mcL_{\text{NLL}}(\theta) =  -  \sum_{k=1}^{|\mcV|} \mathbbm{1}\{y=k\} \log p(y=k \given x;\theta)
\end{equation*}
where $\mathbbm{1}\{\cdot\}$ is the indicator function and $p$ the
distribution from our model (parameterized by $\theta$). This objective can be seen
as minimizing the cross-entropy between the degenerate data
distribution (which has all of its probability mass on one class) and
the model distribution $p(y \given x;\theta)$.

In knowledge distillation, we assume access to a learned teacher distribution $q(y\given x; \theta_T)$,
possibly trained over the same data set. Instead of minimizing cross-entropy with the 
observed data, we instead minimize the cross-entropy with the teacher's probability distribution,
\begin{align*}
\mcL_{\text{KD}}(\theta;\theta_T) =& - \sum_{k=1}^{|\mcV|}  q(y = k \given x;\theta_T) \times \\
& \log p(y = k \given x;\theta)
\end{align*}
where $\theta_T$ parameterizes the teacher distribution and remains fixed.
Note the cross-entropy setup is identical, but the target distribution is no longer a sparse distribution.\footnote{
In some cases the entropy of the teacher/student 
distribution is increased by annealing it with
a temperature term $\tau > 1$
\begin{equation*}
\tilde{p}(y \given  x) \propto p(y \given x)^{\frac{1}{\tau}}
\end{equation*} After testing $\tau \in \{1, 1.5, 2\}$ we
found that  $\tau =1$ worked best.} Training on $q(y \given x;\theta_T)$ is attractive
since it gives more information about other classes for a given data point (e.g.
similarity between classes) and has less variance in gradients \cite{Hinton2015}.

Since this new objective has no direct term for the training data,
it is common practice to interpolate between the two losses,
\begin{equation*}
\mcL(\theta;\theta_T) = (1-\alpha) \mcL_{\text{NLL}}(\theta) + \alpha \mcL_{\text{KD}}(\theta;\theta_T)
\end{equation*}
where $\alpha$ is mixture parameter combining the one-hot distribution and
the teacher distribution.

\section{Knowledge Distillation for NMT} \label{seq-KD}

The large sizes of neural machine translation systems make them  an ideal candidate for 
knowledge distillation approaches. In this section we explore three different 
ways this technique can be applied to NMT.

\subsection{Word-Level Knowledge Distillation}

NMT systems are trained directly to minimize word NLL,
$\mcL_{\text{WORD-NLL}}$, at each position. Therefore 
if we have a teacher model, standard knowledge distillation 
for multi-class cross-entropy can be applied.
We define this distillation for a sentence as,
\begin{eqnarray*}
\mcL_{\text{WORD-KD}} = -\sum_{j=1}^{J} \sum_{k=1}^{|\mcV|} &q(t_{j}=k \given \svec, \tvec_{<j}) \times \\
&\log p(t_{j}=k \given \svec, \tvec_{<j})
\end{eqnarray*}
where $\mcV$ is the target vocabulary set. The student can further be trained to  optimize the
mixture of $\mcL_{\text{WORD-KD}}$ and $\mcL_{\text{WORD-NLL}}$.  In
the context of NMT, we refer to this approach as \textit{word-level}
knowledge distillation and illustrate this in Figure 1 (left).

\subsection{Sequence-Level Knowledge Distillation}

Word-level knowledge distillation allows transfer of these local word
distributions. Ideally however, we would like the student model to mimic the
teacher's actions at the \textit{sequence-level}.  The sequence
distribution is particularly important for NMT, because wrong
predictions can propagate forward at test-time.

First, consider the sequence-level distribution specified by the model over all
possible sequences $\tvec \in \mcT$,
\begin{equation*}
p(\tvec \given \svec) = \prod_{j=1}^J p(t_j \given \svec, \tvec_{<j})
\end{equation*}
for any length $J$.
The sequence-level negative log-likelihood for NMT then involves matching
the one-hot distribution over all complete sequences,
\begin{eqnarray*}
&&\mcL_{\text{SEQ-NLL}} = -\sum_{\tvec \in \mcT} \mathbbm{1}\{\tvec=\yvec\} \log p(\tvec \given \svec) \\ 
&= & -\sum_{j=1}^J \sum_{k=1}^{|\mcV|} \mathbbm{1}\{y_j=k\} \log p(t_{j}=k \given \svec, \tvec_{<j}) 
\\ 
&=& \mcL_{\text{WORD-NLL}}
\end{eqnarray*}
where $\yvec = [y_1, \dots, y_J]$ is the observed sequence.
Of course, this just shows that from a negative log likelihood perspective,
 minimizing word-level NLL and sequence-level NLL are equivalent in this model.

But now consider the case of sequence-level knowledge distillation.  
As before, we
can simply replace the distribution from the data with a probability
distribution derived from our teacher model.
However, instead of using
a single word prediction, we use $q(\tvec \given \svec)$ to represent
the teacher's sequence distribution over the sample space of
all possible sequences,
\begin{equation*}
\mcL_{\text{SEQ-KD}} = -\sum_{\tvec \in \mcT} q(\tvec \given \svec) \log p (\tvec \given \svec)
\end{equation*}
Note that $\mcL_\text{SEQ-KD}$ is inherently different from $\mcL_\text{WORD-KD}$, as the sum is over an
exponential  number of terms. Despite its intractability, we posit that this sequence-level
objective is worthwhile. It gives the teacher the chance to
assign probabilities to complete sequences and therefore transfer a
broader range of knowledge. We thus consider an approximation of this
objective.

Our simplest approximation is to replace the teacher
distribution $q$ with its mode,
\begin{equation*}
q(\tvec \given \svec) \sim \mathbbm{1}\{\tvec = \argmax_{\tvec \in \mcT} q(\tvec \given \svec )\}
\end{equation*}
Observing that finding the mode is itself intractable, we use beam search 
to find an approximation. The loss is then
\begin{eqnarray*}
\mcL_{\text{SEQ-KD}} &\approx&  - \sum_{\tvec \in \mcT} \mathbbm{1}\{\tvec = \hat{\yvec} \} \log p (\tvec \given \svec) \\
&=& - \log p (\tvec = \hat{\yvec} \given \svec)
\end{eqnarray*}
where $\hat{\yvec}$ is now the output from running beam search with the teacher model.

Using the mode seems like a poor approximation for
the teacher distribution $q(\tvec \given \svec)$, as we are
approximating an exponentially-sized distribution with a single
sample. However, previous results showing the effectiveness of
beam search decoding for NMT lead us to belief that a large portion of
$q$'s mass lies in a single output sequence. In fact, in
experiments we find that with beam of size $1$, $q(\hat{\yvec} \given \svec)$ (on average) accounts for
$1.3\%$ of the distribution for German $\rightarrow$ English, and
$2.3\%$ for Thai $\rightarrow$ English (Table 1: $p(\tvec = \hat{\yvec})$).\footnote{Additionally there
  are simple ways to better approximate $q(\tvec \given \svec)$.  One way would be to
  consider a $K$-best list from beam search and renormalizing the
  probabilities,
\begin{equation*}
q(\tvec \given \svec) \sim \frac{q( \tvec \given \svec)}{\sum_{\tvec \in \mathcal{T}_K} q(\tvec \given \svec)}
\end{equation*}
where $\mathcal{T}_K$ is the $K$-best list from beam search. This would increase the training
set by a factor of $K$.  A beam of size
 $5$ captures $2.8\%$ of the distribution for German $\rightarrow$ English, and
 $3.8\%$ for Thai $\rightarrow$ English.
 Another alternative is to use a Monte Carlo estimate and sample from the teacher model
(since
$\mcL_{\text{SEQ-KD}} = \mathbb{E}_{\tvec\sim q(\tvec \given \svec)}[\,-\log p(\tvec \given \svec) \,]$).
However in practice we found the (approximate) mode to work well.}

To summarize, sequence-level knowledge distillation suggests to: (1)
train a teacher model, (2) run beam search over the training set with
this model, (3) train the student network with cross-entropy on this
new dataset. Step (3) is identical to the word-level NLL process
except now on the newly-generated data set. This is shown in Figure 1 (center).

\subsection{Sequence-Level Interpolation}\label{local}

Next we consider integrating the training data back into the process,
such that we train the student model as a mixture of our
sequence-level teacher-generated data ($\mcL_\text{SEQ-KD}$) with the original training data 
($\mcL_\text{SEQ-NLL}$),
\begin{align*}
\mcL &= (1-\alpha)\mcL_{\text{SEQ-NLL}} + \alpha \mcL_{\text{SEQ-KD}} \\
&= -(1-\alpha)\log p(\yvec \given \svec) -\alpha \sum_{\tvec \in \mcT} q(\tvec \given \svec) \log p (\tvec \given \svec)
\end{align*}
where $\yvec$ is the gold target sequence.

Since the second term is intractable, we could again apply the mode
approximation from the previous section,
\begin{equation*}
\mcL = -(1-\alpha)\log p(\yvec \given \svec)  - \alpha \log p (\hat{\yvec} \given \svec)
\end{equation*}
and train on both observed ($\yvec$) and teacher-generated ($\hat{\yvec}$) data.
However, this process is non-ideal for two reasons: (1) unlike for
standard knowledge distribution, it doubles the size of the training
data, and (2) it requires training on both the
teacher-generated sequence and the true sequence, conditioned on the
same source input. The latter concern is particularly problematic since we observe 
that $\yvec$ and $\hat{\yvec}$ are often quite different.   

As an alternative, we propose a single-sequence approximation that
is more attractive in this setting. This approach is inspired by
\textit{local updating} \cite{Liang2006}, a method
 for discriminative training in statistical
machine translation (although to our knowledge not for knowledge
distillation). Local updating suggests selecting a training sequence which is
close to $\yvec$ \textit{and} has high probability under the teacher
model,
\begin{equation*}
\tilde{\yvec} = \argmax_{\tvec \in \mcT} sim(\tvec, \yvec) q(\tvec \given \svec)
\end{equation*}
where $sim$ is a function measuring closeness (e.g. Jaccard
similarity or  BLEU \cite{Papineni2002}). Following local updating, we can approximate this 
sequence by running beam search and choosing
\begin{equation*}
\tilde{\yvec} \approx \argmax_{\tvec \in \mcT_K} sim(\tvec, \yvec)
\end{equation*}
where $\mcT_K$ is the $K$-best list from beam search. We take $sim$ to be smoothed
sentence-level BLEU \cite{Chen2014}.

We justify training on $\tilde{\yvec}$ from a knowledge distillation perspective with the following
generative process: suppose that there is a true target sequence (which we do not observe) that
is first generated from the underlying data distribution  $\mathcal{D}$. And further suppose 
that the target sequence that we observe ($\yvec$) is a noisy version of the unobserved true sequence: i.e. 
(i) $\tvec \sim \mathcal{D}$, (ii) $\yvec \sim \epsilon(\tvec)$, 
where  $\epsilon(\tvec)$ is, for example, a noise function that independently
replaces each element in $\tvec$ with 
a random element in $\mcV$ with some small probability.\footnote{While we employ a simple
  (unrealistic) noise function 
for illustrative purposes, the generative story is quite plausible if we consider a more
elaborate noise function which includes additional sources of noise such as phrase reordering,
replacement of words with synonyms, etc. 
One could view translation having two sources of variance that should be modeled separately: 
variance due to the source sentence ($\tvec \sim \mathcal{D}$), and 
variance due to the individual translator ($\yvec \sim \epsilon(\tvec)$).}
In such a case, ideally the student's
distribution should match the mixture distribution, 
\begin{equation*}
\mathcal{D}_{\text{SEQ-Inter}} \sim (1-\alpha)\mathcal{D} + \alpha q(\tvec \given \svec)
\end{equation*}

In this setting, due to the noise assumption, $\mathcal{D}$ now has significant probability mass around
a neighborhood of $\yvec$ (not just at $\yvec$), and therefore the
$\argmax$ of the  mixture distribution is likely something other
than $\yvec$ (the observed sequence) or $\hat{\yvec}$ (the output from beam search). 
We can see that 
$\tilde{\yvec}$ is a natural approximation to the $\argmax$ of this
mixture distribution between $\mathcal{D}$ and $q(\tvec \given \svec)$
for some $\alpha$. We illustrate this framework in Figure 1 (right) and visualize the distribution
over a real example in Figure 2.

\begin{figure}[t]\label{fig2}
\centering
\includegraphics[width=8cm]{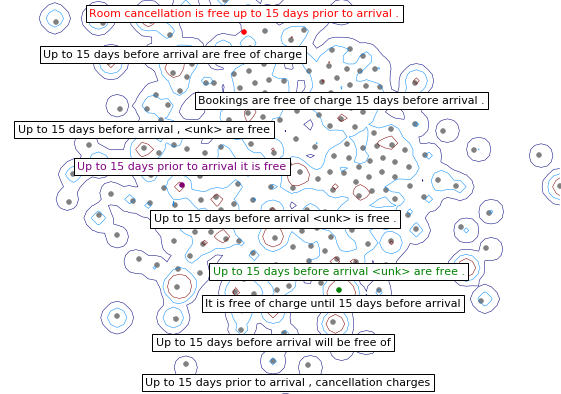}
\caption{Visualization of sequence-level interpolation on an example 
German $\rightarrow$ English sentence:
\textsf{Bis 15 Tage vor Anreise sind Zimmer-Annullationen kostenlos}.
We run beam search, plot the final hidden state
of the hypotheses using t-SNE 
and show the corresponding (smoothed)
probabilities with contours. 
In the above example, the sentence that is at the top of the beam after beam search 
(green) is quite far away from gold (red), so we train the model on a sentence
that is on the beam but had the highest $sim$ (e.g. BLEU) to gold (purple).}
\end{figure}

\section{Experimental Setup}
To test out these approaches, 
we conduct two sets of NMT experiments:  high resource (English $\rightarrow$ German) and low resource
(Thai $\rightarrow$ English).

The \textbf{English-German} data comes from WMT 2014.\footnote{http://statmt.org/wmt14}
The training set has $4$m sentences and we take  newstest2012/newstest2013
as the dev set and newstest2014 as the test set. We keep the top $50$k most
frequent words, and replace the rest with UNK. The teacher model
is a $4 \times 1000$ LSTM (as in \newcite{Luong2015}) and we train two student models:
$2 \times 300$ and $2 \times 500$.
The \textbf{Thai-English} data comes from IWSLT 2015.\footnote{https://sites.google.com/site/iwsltevaluation2015/mt-track}
There are $90$k sentences in the training set and we take 2010/2011/2012 data as the dev set and
2012/2013 as the test set, with a vocabulary size is $25$k. Size of the teacher model is $2 \times 500$ 
(which performed better than $4 \times 1000$, $2 \times 750$ models), and the student model is $2 \times 100$. 
Other training details mirror \newcite{Luong2015}.

We evaluate on tokenized BLEU with  $\texttt{multi-bleu.perl}$, and experiment with the following variations:
\paragraph{Word-Level Knowledge Distillation (Word-KD)} Student is 
trained on the original data and additionally trained to minimize the cross-entropy of
the teacher distribution at the word-level. We tested $\alpha \in \{0.5, 0.9\}$ and found
$\alpha = 0.5$ to work better.
 \paragraph{Sequence-Level Knowledge Distillation (Seq-KD)} Student is trained on
  the teacher-generated data, which is the result of running beam search
  and taking the highest-scoring sequence with the teacher model. We use beam size $K= 5$ (we did not see
improvements with a larger beam).
 \paragraph{Sequence-Level Interpolation (Seq-Inter)} Student is trained on the sequence on
  the teacher's beam that had the highest BLEU (beam size $K=35$). We adopt a fine-tuning approach where we begin training from a pretrained model 
(either on original data or Seq-KD data) and train with a smaller learning rate ($0.1$). 
 For English-German we generate Seq-Inter data on a smaller portion of the
 training set  ($\sim50\%$) for efficiency.

The above methods are complementary and can be combined with each other. For example,
we can train on teacher-generated data but still include a word-level cross-entropy term
between the teacher/student (Seq-KD $+$ Word-KD in Table 1), or fine-tune towards
Seq-Inter data starting from the baseline model trained on original data
(Baseline $+$ Seq-Inter in Table 1).\footnote{For instance, `Seq-KD $+$ Seq-Inter $+$ Word-KD' in Table 1 means
that the model was trained on Seq-KD data and fine-tuned towards Seq-Inter data with the
mixture cross-entropy loss at the word-level.}

\section{Results and Discussion}\label{results}
\begin{table*}[!ht] \label{de}
\centering
\small
\begin{tabular}{l c c  c c r r r }
\toprule
Model &    BLEU$_{K=1}$ & $\Delta_{K=1}$ & BLEU$_{K=5}$ & $\Delta_{K=5}$ & & PPL & $p(\tvec = \hat{\yvec})$ \\
\midrule
\textit{English $\rightarrow$ German WMT 2014} \\ 
\midrule
Teacher Baseline $4 \times 1000$  (Params: $221$m)  & $17.7$ &  $-$ & $19.5$&   $-$ & &  $6.7$ &  $1.3\%$ \\
\hspace{4mm} Baseline $+$  Seq-Inter  & $19.6$ & $+1.9$&  $19.8$& $+0.3$&  & $10.4$ & $8.2\%$   \\
\midrule
Student Baseline $2 \times 500$ $\,$ (Params: $84$m)  & $14.7$ & $-$ & $17.6$&  $-$ & & $8.2$ & $0.9\%$  \\
\hspace{4mm} Word-KD  & $15.4$ & $+0.7$& $17.7$& $+0.1$& & $8.0$ & $1.0\%$  \\
\hspace{4mm} Seq-KD   & $18.9$ & $+\mathbf{4.2}$& $19.0$& $+1.4$& & $22.7$ & $16.9\%$ \\
\hspace{4mm} Baseline $+$ Seq-Inter   & $18.5$ & $+3.6$& $18.7$& $+1.1$& & $11.3$ & $5.7\%$ \\
\hspace{4mm} Word-KD $+$ Seq-Inter  & $18.3$ & $+3.6$& $18.5$& $+0.9$& & $11.8$ & $6.3\%$ \\
\hspace{4mm} Seq-KD $+$ Seq-Inter  & $18.9$ & $+\mathbf{4.2}$&$19.3$ & $+\mathbf{1.7}$ & & $15.8$ & $7.6\%$  \\
\hspace{4mm} Seq-KD $+$ Word-KD  & $18.7$ & $+4.0$& $18.9$& $+1.3$& & $10.9$ & $4.1\%$  \\
\hspace{4mm} Seq-KD $+$ Seq-Inter $+$ Word-KD  & $18.8$ & $+4.1$& $19.2$& $+1.6$& & $14.8$ &$7.1\%$  \\
\midrule
Student Baseline $2 \times 300$ $\,$ (Params: $49$m) & $14.1$ & $-$ & $16.9$&  $-$ &  & $10.3$ & $0.6\%$ \\
\hspace{4mm}  Word-KD  & $14.9$ & $+0.8$& $17.6$& $+0.7$& & $10.9$ & $0.7\%$  \\
\hspace{4mm}  Seq-KD  & $18.1$ & $+4.0$& $18.1$& $+1.2$&  & $64.4$ & $14.8\%$  \\
 \hspace{4mm} Baseline $+$ Seq-Inter  & $17.6$ & $+3.5$& $17.9$& $+1.0$&  & $13.0$ & $10.0\%$ \\
\hspace{4mm} Word-KD $+$ Seq-Inter   & $17.8$ & $+3.7$& $18.0$& $+1.1$& & $14.5$ & $4.3\%$ \\
\hspace{4mm}  Seq-KD $+$ Seq-Inter  & $18.2$ & $+4.1$& $18.5$& $+1.6$&  & $40.8$ & $5.6\%$ \\
\hspace{4mm} Seq-KD $+$ Word-KD & $17.9$ & $+3.8$& $18.8$& $+1.9$& & $44.1$ & $3.1\%$  \\
\hspace{4mm} Seq-KD $+$ Seq-Inter $+$ Word-KD  & $18.5$ & $+\mathbf{4.4}$& $18.9$& $+\mathbf{2.0}$& & $97.1$& $5.9\%$  \\

\bottomrule
\toprule
\textit{Thai $\rightarrow$ English IWSLT 2015} \\ 
\midrule
Teacher Baseline $2 \times 500$ $\,$ (Params: $47$m) & $14.3$ &  $-$ & $15.7$&   $-$ &  & $22.9$ & $2.3\%$ \\
\hspace{4mm}  Baseline $+$ Seq-Inter   & $15.6$ & $+1.3$& $16.0$& $+0.3$& & $55.1$ &  $6.8\%$ \\
\midrule
Student Baseline $2 \times 100$ $\,$ (Params: $8$m) & $10.6$ & $-$ & $12.7$&   $-$ & & $37.0$ & $1.4\%$  \\
\hspace{4mm}  Word-KD   & $11.8$ & $+1.2$& $13.6$&  $+0.9$& & $35.3$ &  $1.4\%$\\
\hspace{4mm}  Seq-KD   & $12.8$ & $+2.2$& $13.4$& $+0.7$& & $125.4$ & $6.9\%$  \\
\hspace{4mm}  Baseline $+$ Seq-Inter   & $12.9$ & $+2.3$& $13.1$& $+0.4$& & $52.8$ & $2.5\%$  \\
\hspace{4mm} Word-KD $+$ Seq-Inter  & $13.0$ & $+2.4$& $13.7$& $+1.0$ & & $58.7$ &  $3.2\%$  \\
\hspace{4mm}  Seq-KD $+$ Seq-Inter   & $13.6$ & $+3.0$& $14.0$&  $+1.3$& & $106.4$ & $3.9\%$  \\
\hspace{4mm}  Seq-KD $+$ Word-KD   & $13.7$ & $+3.1 $& $14.2$&  $+1.5$ && $67.4$& $3.1\%$ \\
\hspace{4mm} Seq-KD $+$ Seq-Inter $+$ Word-KD   & $14.2$ & $+\mathbf{3.6}$& $14.4$& $+\mathbf{1.7}$ & & $117.4$ & $3.2\%$ \\
\bottomrule
\end{tabular}
\caption{Results on English-German (newstest2014) and Thai-English (2012/2013) test sets.
BLEU$_{K=1}$: BLEU score with beam size $K=1$ (i.e. greedy
decoding); $\Delta_{K=1}$: BLEU gain over the baseline model without any knowledge distillation with greedy decoding;  BLEU$_{K=5}$: BLEU score with beam size $K=5$;
 $\Delta_{K=5}$: BLEU gain over the baseline model without any knowledge distillation with beam size $K = 5$; 
PPL: perplexity on the test set; $p(\tvec = \hat{\yvec})$: Probability of output sequence from greedy decoding 
(averaged over the test set). Params: number of parameters in the model. Best results (as measured by improvement over the baseline)
within each category are highlighted in bold.}
\end{table*}

Results of our experiments are shown in Table 1. We find that while
word-level knowledge distillation (Word-KD) does improve upon the
baseline, sequence-level knowledge distillation (Seq-KD) does better
on English $\rightarrow$ German and performs similarly on Thai
$\rightarrow$ English. Combining them (Seq-KD $+$ Word-KD) results in further gains
for the $2\times 300$ and $2\times100$ models (although not for the $2\times500$ model),
indicating that these methods provide orthogonal means of transferring knowledge
from the teacher to the student: Word-KD is transferring knowledge at the
 the local (i.e. word) level while Seq-KD is transferring knowledge
at the global (i.e. sequence) level.

Sequence-level interpolation (Seq-Inter), in addition to improving models trained via
Word-KD and Seq-KD, 
also improves upon the original teacher model that was trained on the actual data
but fine-tuned towards Seq-Inter data (Baseline $+$ Seq-Inter). 
In fact, greedy decoding with this fine-tuned model has similar
performance ($19.6$) as beam search with the original model ($19.5$), allowing for faster
decoding even with an identically-sized model. 

We hypothesize that sequence-level knowledge distillation is effective because it allows the student network to only model
relevant parts of the teacher distribution (i.e. around the teacher's mode) instead of `wasting' parameters on trying to
model the entire space of translations.
Our results suggest that this is indeed the case: the probability mass that Seq-KD models assign to the approximate mode 
is much higher than is the case for baseline models trained on original data (Table 1: $p(\tvec = \hat{\yvec})$). 
For example, on English $\rightarrow$ German the (approximate) $\argmax$ for the $2 \times 500$ Seq-KD model 
(on average) accounts for $16.9\%$ of the total probability mass, while the corresponding number is $0.9\%$ for the baseline. 
This also explains the success of 
greedy decoding for Seq-KD models---since we are only modeling around the teacher's mode, the student's distribution
is more peaked and therefore the $\argmax$ is much easier to find. Seq-Inter offers a compromise between the two, with
the greedily-decoded sequence accounting for $7.6\%$ of the distribution.

Finally, although past work has shown that models with lower perplexity generally tend to have higher BLEU,
our results indicate that this is not necessarily the case. The perplexity of the baseline $2\times500$  English $\rightarrow$ German 
 model is $8.2$ while the perplexity of the 
corresponding Seq-KD model is $22.7$, despite the fact that Seq-KD model does significantly better for both greedy ($+4.2$ BLEU)
and beam search ($+1.4$ BLEU) decoding. 


\subsection{Decoding Speed}
\begin{table}[t] \label{speed}
\centering
\small
\begin{tabular}{l  r  r  r  }
\toprule
Model Size  & GPU & CPU & Android \\
\midrule
\textit{Beam = 1 (Greedy)}\\
\midrule 
$4 \times 1000$ &$425.5$& $15.0$& $-$   \\
$2 \times 500$ &$1051.3$& $63.6$&$8.8$   \\
$2 \times 300$& $1267.8$& $104.3$&$15.8$   \\
\midrule
\textit{Beam $=5$} \\
\midrule
$4 \times 1000$ &$101.9$ & $7.9$ & $-$  \\
$2 \times 500$ & $181.9$&  $22.1$ & $1.9$ \\
$2 \times 300$ & $189.1$&  $38.4$ &$3.4$ \\
\bottomrule
\end{tabular}
\caption{Number of source words translated per second across GPU (GeForce GTX Titan X), 
CPU, and smartphone (Samsung Galaxy 6) for the various English $\rightarrow$ German models. 
We were unable to open the $4 \times 1000$ model on the smartphone.}
\end{table}

Run-time complexity for beam search grows linearly with beam size. Therefore,
the fact that sequence-level knowledge distillation allows for greedy decoding  is 
 significant, with practical implications for running NMT systems across various devices. 
 To test the  speed gains,
we run the teacher/student models on GPU, CPU, and smartphone, and check the average
number of source words translated per second (Table 2). We use a GeForce GTX Titan X for 
GPU and a Samsung Galaxy 6 smartphone. We find that we can run the
student model $10$ times faster with greedy decoding than the teacher model with beam search 
on GPU ($1051.3$ vs $101.9$ words/sec), with similar performance.

\subsection{Weight Pruning}

\begin{table}[t] \label{prune}
\centering
\small
\begin{tabular}{l  r  r c  r }
\toprule
Model & Prune $\%$ & Params & BLEU & Ratio \\
\midrule 
$4 \times 1000$ & $0\%$ &$221$ m& $19.5$& $1 \times$   \\
$2 \times 500$ &  $0\%$& $84$ m& $19.3$& $3 \times$   \\
\midrule
$2 \times 500$ & $50\%$& $42$ m&  $19.3$ & $5 \times$ \\
$2 \times 500$ &  $80\%$& $17$ m&  $19.1$ & $13 \times$ \\
$2 \times 500$ &  $85\%$& $13$ m&  $18.8$ & $18 \times$ \\
$2 \times 500$ &  $90\%$& $8$ m &  $18.5$  & $26 \times$ \\

\bottomrule
\end{tabular}
\caption{Performance of student models with varying $\%$ of the weights pruned. Top two
rows are models without any pruning.
Params: number of parameters in the model; Prune $\%$: Percentage of weights pruned based on their absolute values;
BLEU: BLEU score with beam search decoding ($K = 5$) after retraining the pruned model; 
Ratio: Ratio of the number of parameters versus the original teacher model 
(which has $221$m parameters).
}
\end{table}

Although knowledge distillation enables training faster models,
the number of parameters for the student models is still somewhat
large (Table 1: Params), due to the word embeddings which dominate most of the parameters.\footnote{Word
embeddings scale linearly while RNN parameters scale quadratically with the dimension size.}
For example, on the $2 \times 500$ English $\rightarrow$ German model
the word embeddings account for approximately $63\%$ ($50$m out of $84$m) of the parameters.
The size of word embeddings have little impact on run-time as the word embedding layer is
 a simple lookup table that only affects the first layer of the model.

We therefore focus next on reducing the memory footprint of the student models further through 
weight pruning. Weight pruning for NMT was recently investigated by \newcite{See2016}, who found that up to 
$80-90\%$ of the parameters in a large NMT model can be pruned with little loss in performance.
We take our best 
English $\rightarrow$ German student model ($2 \times 500$ Seq-KD $+$ Seq-Inter) and prune $x\%$ of the parameters by removing
the weights with the lowest absolute values.
We then retrain the pruned model on Seq-KD data with a learning rate of $0.2$ 
and fine-tune towards Seq-Inter data with a learning rate of $0.1$. As observed by \newcite{See2016},
retraining proved to be crucial. The results are shown in Table 3. 

Our findings suggest that compression benefits achieved through 
weight pruning and knowledge distillation are 
orthogonal.\footnote{To our knowledge combining pruning and knowledge distillation has not been investigated before.}
 Pruning $80\%$
 of the weight in the $2 \times 500$ student model results in a model with $13\times$ fewer parameters than the original teacher 
model with only a decrease of $0.4$ BLEU.
 While pruning $90\%$ of the weights results in a more appreciable decrease of $1.0$ BLEU, 
the model is drastically smaller with $8$m parameters, which is $26\times$ fewer than the original teacher model.

\subsection{Further Observations}
\begin{itemize}
\item For models trained with word-level knowledge distillation, we also tried regressing the student network's
top-most hidden layer at each time step to the teacher network's top-most hidden layer as
a pretraining step, noting that \newcite{Romero2015} obtained improvements with a 
similar technique on feed-forward models.
We found this to give comparable results to standard knowledge distillation and 
hence did not pursue this further.
\item There have been promising recent results
on eliminating word embeddings completely
and obtaining word representations directly from characters 
with character composition models, which have many fewer parameters than word embedding
lookup tables
\cite{Ling2015,Kim2016,Ling2015b,Jozefowicz2016,Jussa2016}. 
Combining such methods with knowledge distillation/pruning 
to further reduce the memory footprint of
NMT systems remains an avenue for future work.
\end{itemize}
\section{Related Work}
Compressing deep learning models is an active area of current research.
\textit{Pruning} methods involve pruning weights or entire neurons/nodes based on
some criterion. \newcite{LeCun1990} prune weights
based on an approximation of the Hessian, while \newcite{Han2016} show that a simple
magnitude-based pruning works well. Prior work on removing neurons/nodes include \newcite{Srinivas2015} 
and \newcite{Mariet2016}. \newcite{See2016} were the first to apply pruning  
 to Neural Machine Translation,
observing that that different parts of the architecture 
(input word embeddings, LSTM matrices, etc.) admit different
levels of pruning.
\textit{Knowledge distillation} approaches train a smaller student model to mimic a larger teacher model, by minimizing 
the loss between the teacher/student predictions \cite{Bucila2006,Ba2014,Li2014,Hinton2015}.
\newcite{Romero2015} additionally regress on the intermediate hidden layers of the student/teacher
network as a pretraining step, while \newcite{Mou2015} obtain smaller word embeddings from a teacher model via regression.
There has also been work on transferring knowledge across different network 
architectures:  \newcite{Chan2015b} show that a 
deep non-recurrent neural network can learn from an RNN; \newcite{Geras2016}
train a CNN to mimic an LSTM for speech recognition. \newcite{Kuncoro2016} recently investigated
knowledge distillation for structured prediction by having a single parser learn
from an ensemble of parsers.

Other approaches for compression involve low rank factorizations of weight matrices \cite{Denton2014,Jaderberg2014,Lu2016,Prabhavalkar2016},
sparsity-inducing regularizers \cite{Murray2015}, binarization of weights \cite{Courbariaux2016,Lin2016},
and weight sharing \cite{Chen2015b,Han2016}. 
Finally, although we have motivated sequence-level knowledge distillation in the context of training a smaller model,
there are other techniques that train on a mixture of the model's predictions and the data, such as 
local updating \cite{Liang2006}, hope/fear training \cite{Chiang2012}, SEARN \cite{Daume2009},
DAgger \cite{Ross2011}, and minimum risk training \cite{Och2003,Shen2016}.

\section{Conclusion}
In this work we have investigated existing knowledge distillation methods for NMT (which work at the
word-level) and
introduced two \textit{sequence-level} variants of knowledge distillation, which provide
improvements over standard word-level knowledge distillation.

We have chosen to focus on translation as this domain has generally required the largest
capacity deep learning models, but the sequence-to-sequence framework has been successfully applied
to a wide range of tasks including parsing \cite{Vinyals2015}, summarization \cite{Rush2015}, 
dialogue \cite{Vinyals2015c,Serban2016,Li2016}, NER/POS-tagging \cite{Gillick2016},
image captioning \cite{Vinyals2015b,Xu2015}, video generation \cite{Srivastava2015a},
and speech recognition \cite{Chan2015}.
We anticipate that methods described in this paper can be used to similarly train smaller
models in other domains. 

\bibliography{master}
\bibliographystyle{emnlp2016}

\end{document}